%% file: main.tex
\pdfoutput=1

\documentclass[11pt]{article}

\usepackage{acl}

\usepackage{times}
\usepackage{latexsym}

\usepackage[T1]{fontenc}

\usepackage[utf8]{inputenc}

\usepackage{microtype}

\usepackage{inconsolata}

\input{package.tex}

\title{From Instructions to Constraints: \\ Language Model Alignment with Automatic Constraint Verification}

\author{
\textbf{Fei Wang}$^{\dagger*}$ \quad 
\textbf{Chao Shang}$^{\ddagger}$ \quad 
\textbf{Sarthak Jain}$^{\ddagger}$ \quad 
\textbf{Shuai Wang}$^{\ddagger}$ \quad 
\textbf{Qiang Ning}$^{\ddagger}$  \\ 
\textbf{Bonan Min}$^{\ddagger}$ \quad 
\textbf{Vittorio Castelli}$^{\ddagger}$ \quad 
\textbf{Yassine Benajiba}$^{\ddagger}$ \quad 
\textbf{Dan Roth}$^{\ddagger}$ \\
$^\dagger$University of Southern California \quad
$^\ddagger$AWS AI Labs \\
\small \texttt{fwang598@usc.edu} \quad \texttt{\{chshang,jsarth,wshui,qning,bonanmin,vittorca,benajiy,drot\}@amazon.com}
}

\begin{document}
\maketitle

\renewcommand{\thefootnote}{\fnsymbol{footnote}}
\footnotetext[1]{Work done during internship at AWS AI Labs.}
\renewcommand{\thefootnote}{\arabic{footnote}}

% \abovedisplayskip=5pt
% \abovedisplayshortskip=1pt
% \belowdisplayskip=5pt
% \belowdisplayshortskip=1pt

\input{section/0_abstract}

\input{section/1_introduction}

\input{section/2_method}
\input{section/3_experiment}

\input{section/5_related_work}

\input{section/6_conclusion}

\bibliography{anthology,custom}
\bibliographystyle{acl_natbib}

\input{section/7_appendix}

\end{document}

%% file: package.tex
\usepackage{refstyle}
\usepackage{amsmath}
\usepackage{cleveref}
\usepackage{enumerate}
\usepackage{xspace}
\usepackage{tikz}
\usepackage{graphicx}
\usepackage{pgf-pie}
\usepackage{pgfplots}
\pgfplotsset{compat=1.17}

\usepackage{tcolorbox}
\usepackage{listings}
\usepackage[frozencache,cachedir=.]{minted}

\usepackage{booktabs}
\usepackage{multirow} %
\usepackage{soul}%
\usepackage{tabularx}

\usepackage{pifont}

\usepackage{graphicx}

\crefformat{section}{\S#2#1#3}
\crefformat{subsection}{\S#2#1#3}
\crefformat{subsubsection}{\S#2#1#3}
\crefrangeformat{section}{\S#3#1#4 to~\S#5#2#6}
\crefmultiformat{section}{\S#2#1#3}{ and~\S#2#1#3}{, #2#1#3}{ and~#2#1#3}
\Crefformat{figure}{#2Fig.~#1#3}
\Crefmultiformat{figure}{Figs.~#2#1#3}{ and~#2#1#3}{, #2#1#3}{ and~#2#1#3}
\Crefformat{table}{#2Tab.~#1#3}
\Crefmultiformat{table}{Tabs.~#2#1#3}{ and~#2#1#3}{, #2#1#3}{ and~#2#1#3}
\Crefformat{appendix}{#2Appx.~\S#1#3}
\Crefmultiformat{appendix}{Appx.~#2#1#3}{ and~#2#1#3}{, #2#1#3}{ and~#2#1#3}
\crefformat{algorithm}{Alg.~#2#1#3}
\Crefformat{equation}{#2Eq.~#1#3}

\newcommand{\stitle}[1]{\vspace{1ex} \noindent{\bf #1.}}

\newcommand{\MODEL}{\mbox{\textsc{ACT}}\xspace}
\newcommand{\MODELFULL}{\mbox{\textsc{ACT}\xspace(\underline{A}ligning to \underline{C}ons\underline{T}raints})}

%% file: section/0_abstract.tex
\begin{abstract}

User alignment is crucial for adapting general-purpose language models (LMs) to downstream tasks, but human annotations are often not available for all types of instructions, especially those with customized constraints.
We observe that user instructions typically contain constraints.
While assessing response quality in terms of the whole instruction is often costly, efficiently evaluating the satisfaction rate of constraints is feasible.
We investigate common constraints in NLP tasks, categorize them into three classes based on the types of their arguments, and propose a unified framework, \MODELFULL, to automatically produce supervision signals for user alignment with constraints. 
Specifically, \MODEL uses constraint verifiers, which are typically easy to implement in practice, to compute constraint satisfaction rate (CSR) of each response.
It samples multiple responses for each prompt and collect preference labels based on their CSR automatically. 
Subsequently, \MODEL adapts the LM to the target task through a ranking-based learning process.
Experiments on fine-grained entity typing, abstractive summarization, and temporal question answering show that \MODEL is able to enhance LMs' capability to adhere to different classes of constraints, thereby improving task performance.
Further experiments show that the constraint-following capabilities are transferable.%

\end{abstract}

%% file: section/1_introduction.tex
\section{Introduction}

User alignment is crucial for adapting general-purpose language models (LMs) to downstream tasks, which typically necessitates the meticulous collection of human annotation to integrate tailored knowledge linked to user instructions \cite{zhang2023instruction,ouyang2022training,mishra-etal-2022-cross}.
However, human annotations are often not available for all types of instructions, especially those with customized constraints.
Recent research indicates that even aligned LMs face challenges in effectively satisfying numerous natural and easily expressible constraints across various common NLP tasks \cite{sun2023evaluating,qin2024infobench,jiang2023followbench,abdin2023kitab}. 
This highlights the difficulty of using limited human annotation in addressing diverse and ever-changing user needs.

\begin{figure}
  \centering
  \includegraphics[width=\columnwidth]{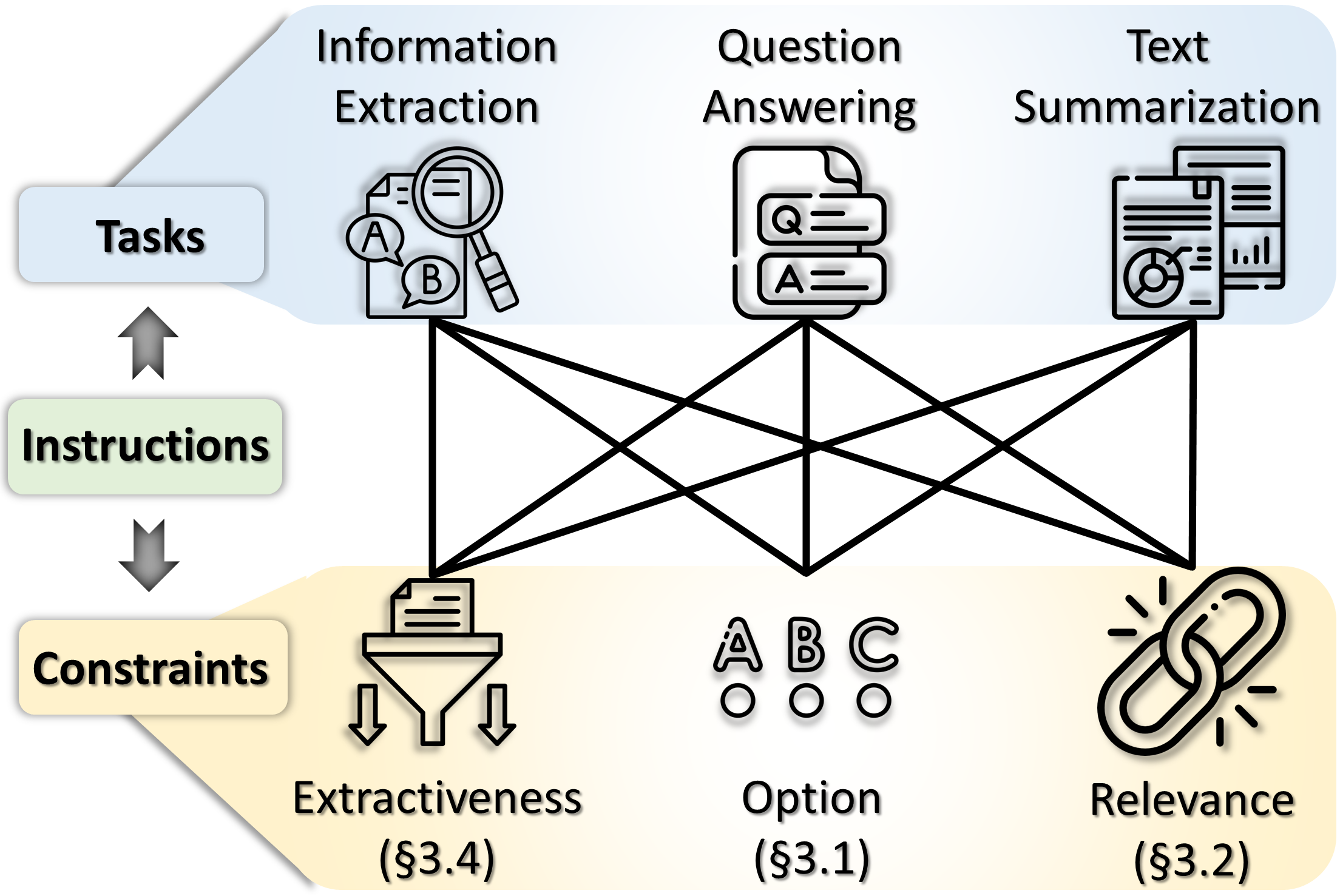}
  \caption{Each user instruction contains one or more constraints. The same task may be associated with different constraints depending on user intents, whereas different tasks may share similar constraints.}
  \label{fig/constraints}
  % \vspace{-1em}
\end{figure}

\begin{figure}
  \centering
  \includegraphics[width=0.95\columnwidth]{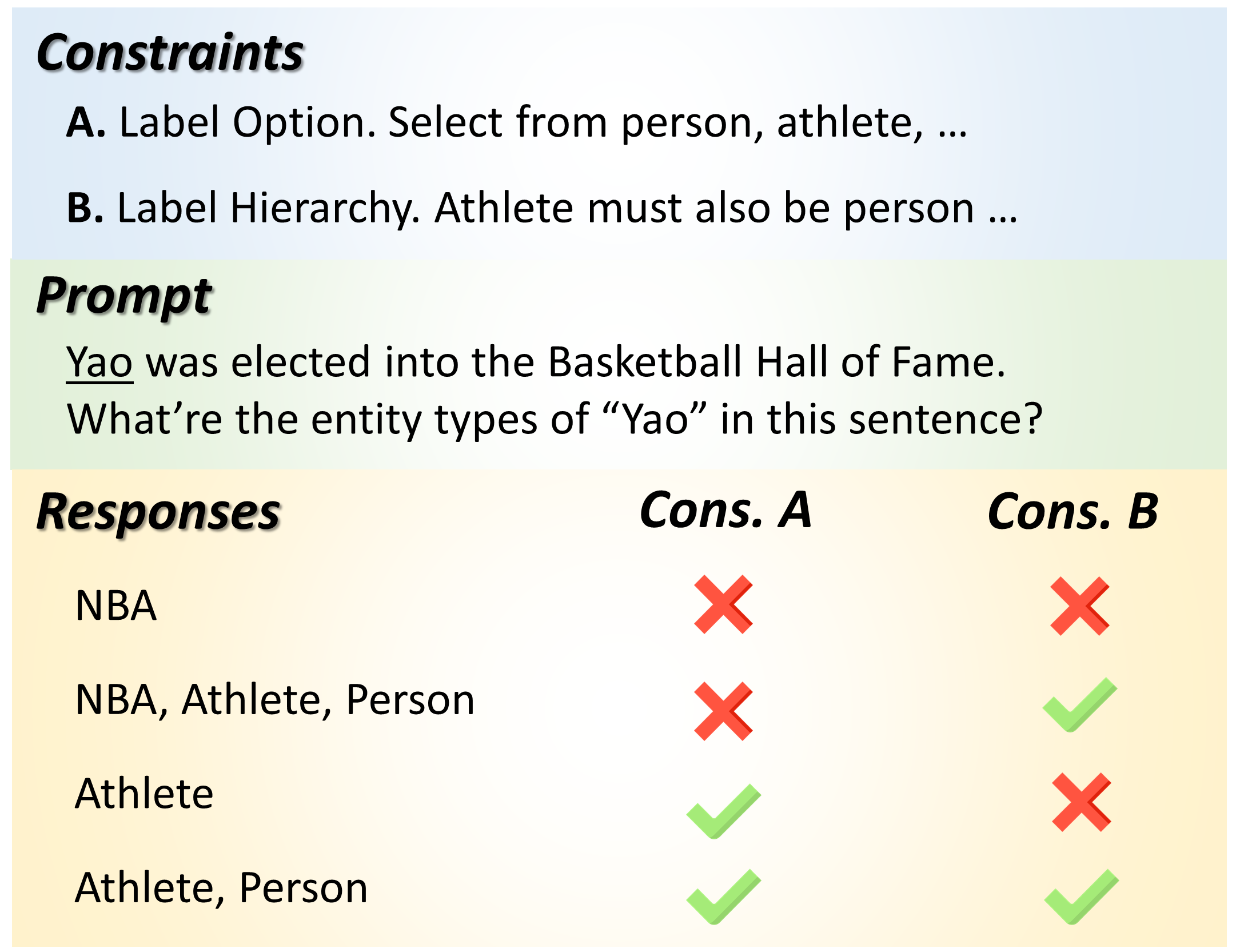}
  \caption{An example of fine-grained entity typing with label option and label hierarchy constraints. A feasible response must satisfy both constraints.}
  \label{fig/example}
  % \vspace{-1em}
\end{figure}

When collecting alignment data, previous work typically treats each instruction (with an instance) as an indivisible whole.
This leads to the necessity of independent annotation for each unique instruction (and each unique instance). The complex process of assessing response quality invariably requires human involvement.\footnote{The situation becomes even more challenging when the user combines various instances with the instruction, as each instance alters the instruction, demanding additional annotation efforts.}
We observe that user instructions typically contain explicit or implicit constraints.
Constraints are generally shared across instances (and tasks) and are much easier to evaluate, thereby facilitating \textit{efficient data annotation}.
As shown in \Cref{fig/constraints}, both information extraction and question answering may have the \texttt{Option} constraint, requiring the LM to select from given options. 
This constraint is universally applicable to all instances in these tasks and can be automatically verified by comparing the model response and the options.
Constraints also reveal crucial insights into user intents, providing valuable information for \textit{effective LM alignment}.
They can help approximate the solution space, identify prediction errors, and guide the model toward the correct answer \cite{chang-etal-2007-guiding,wang2023regularization,ning-etal-2018-joint,wang-etal-2020-joint}.
\Cref{fig/example} illustrates the functionality of constraints in detail. The fine-grained entity typing task has two natural constraints -- label option and label hierarchy describing the relationship of different entity types. 
While obtaining the precise answer needs human annotation, automatic constraint verification can already identify numerous incorrect responses.

In this paper, we investigate common constraints in NLP tasks, categorize them into three classes based on the types of their arguments, and propose a unified LM alignment framework, \MODELFULL, using automatic constraint verifiers to provide supervision signals for adapting models to downstream tasks (\Cref{sec/method}). 
As shown in \Cref{fig/act}, \MODEL starts from selecting constraints that can provide essential knowledge about user intents while at the same time automatically verifiable. 
Then, the constraint verifiers can efficiently measure constraint satisfaction rate (CSR) of model responses.
These verifiers are typically easy to implement and are applicable to all instances governed by the corresponding constraints. 
With their assistance, \MODEL gathers supervision signals for LM alignment from unlabeled instances.
It samples multiple responses for each unlabeled instance and automatically assigns relative preferences to them based on their CSR.
Through a ranking-based learning process \cite{yuan2023rrhf,liu-etal-2022-brio}, \MODEL incorporates the knowledge revealed by the constraints into the LM.

We verify the effectiveness of our method on each class of constraints, taking fine-grained entity typing \cite{ling2012fine}, abstractive text summarization \cite{narayan-etal-2018-dont}, and temporal question answering \cite{ning-etal-2020-torque} as examples (\Cref{sec/experiment}).
Experimental results show that our method, even with little or no labeled data, can significantly enhance model capabilities on downstream tasks, achieving comparable performance to finetuning with the same amount of labeled data.
A pilot study on three different tasks, all sharing the extractiveness constraint, further demonstrates the transferability of learned constraints. 
Our work not only presents a framework for aligning LMs to diverse user instructions (or user-specified tasks) without human annotation, but also suggests the feasibility of tuning LMs with general constraint-following capabilities in a cost-efficient manner.

Our contributions are three-fold.
First, we decompose constraints from instructions, offering efficient data annotation and facilitating effective LM alignment. In this context, we formally define three classes of constraints.
Second, we propose \MODEL, a unified and cost-efficient LM alignment framework for adapting LMs to downstream tasks using the feedback from automatic constraint verifiers.
Third, experimental results on various tasks and constraints show the effectiveness of our method on all classes of constraints and demonstrate the transferability of constraint-following capabilities.

%% file: section/2_method.tex
\section{Method}
\label{sec/method}

\begin{figure*}
  \centering
  \includegraphics[width=0.95\textwidth]{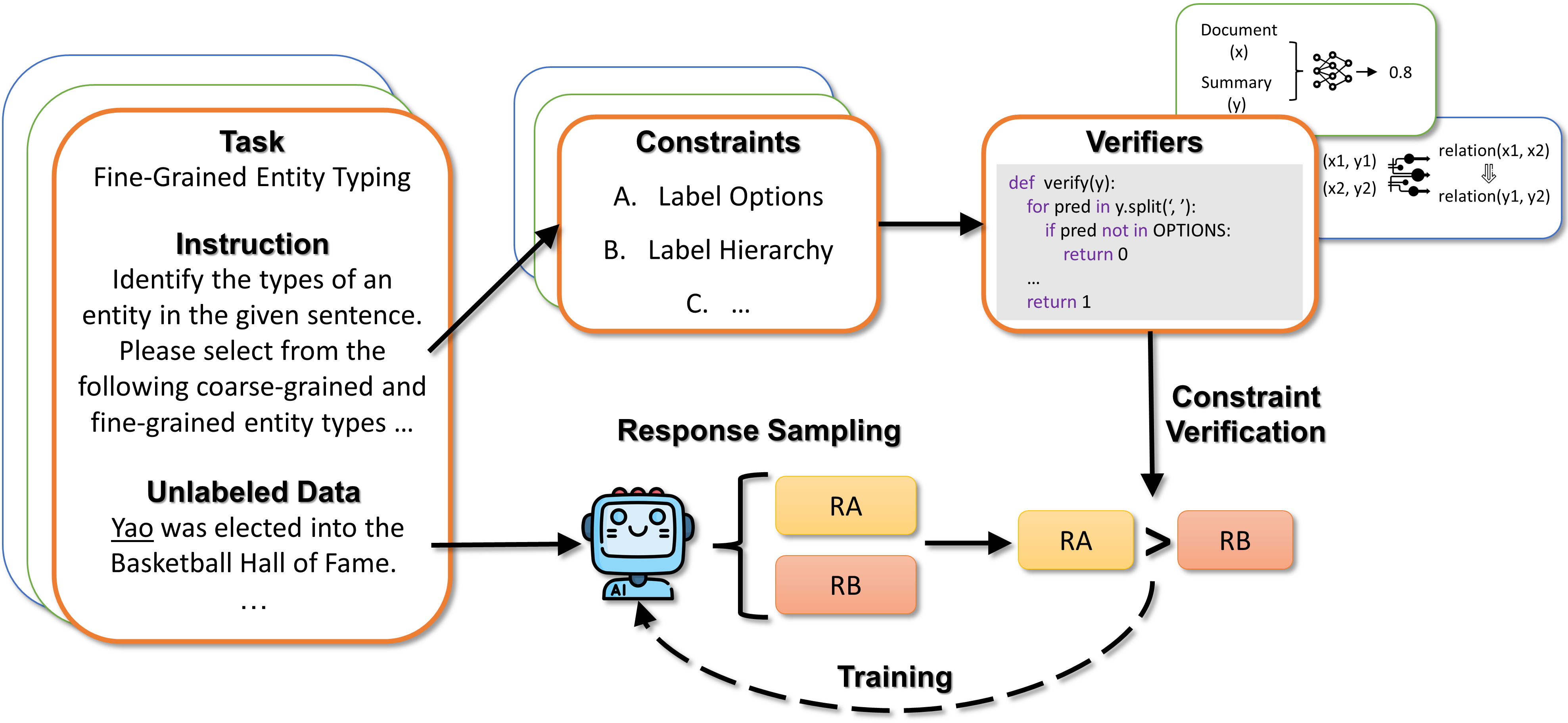}
  \caption{Overview of \MODEL. \MODEL utilizes automatic constraint verifiers, which are typically easy to implement in practice, to assess how well a response satisfies the constraints specified in the instruction. It samples two or more responses (e.g., RA and RB) for each prompt. Then, it computes the constraint satisfaction rate (CSR) of each response and assigns the preference label to each response pair based on their CSR (e.g., RA is better than RB). The preference labels serve as supervision signals for LM alignment.}
  \label{fig/act}
  %\vspace{-1em}
\end{figure*}

We seek to build a unified framework to align LMs with various constraints.
As shown in \Cref{fig/act}, the \MODEL framework starts from selecting proper constraints (\Cref{sec/method/constraint}) and implementing corresponding constraint verifiers (\Cref{sec/method/verifier}).
Then, it samples multiple responses for each instance in the unlabeled task dataset (\Cref{sec/method/response}).
The automatic constraint verifiers will measure the constraint satisfaction rate of responses and provide supervision signals for model alignment (\Cref{sec/method/verification}).
Finally, \MODEL aligns the model with constraints for adaptation (\Cref{sec/method/training}).

\subsection{Constraint Selection}
\label{sec/method/constraint}
Formally, we define constraint as a function $f$ that verifies the satisfiablity of the prompt $x$ and the model response $y$. 
Derived from user instructions, they verify essential requirements for fulfilling user intents.
According to the argument of $f$, we categorize task constraints into three classes: 
\begin{itemize}
    \setlength\itemsep{0em}
    \setlength\itemindent{0em}
    \item $f(y)$ defines a constraint for a response, such as response length, response format, and response candidate. For example, the fine-grained entity typing task requires the LM to respond with given options.
    \item $f(x,y)$ defines a constraint for a prompt-response pair. This type of constraint requires comparing the model input and output, such as their relevance and text overlap. For example, the abstractive summarization task expect a high relevance between the input document and the model-generated summary.
    \item $f(\{x_i,y_i\})$ defines a constraint for multiple prompt-response pairs. This type of constraint involves comparing multiple instances, such as the logical consistency of answers to related questions. For example, in temporal question answering, the answers to \textit{"what happens before event A"} and \textit{"what happens after event A"} should have no overlap. 
\end{itemize}
In \MODEL, constraints should possess two properties: revealing \textit{essential knowledge} and being \textit{automatically verifiable}.
Generally, constraints that more precisely approximate the user intent are more effective in LM alignment. 
\MODEL can combine multiple constraints from different perspectives to achieve a more effective approximation.

\subsection{Verifier Realization}
\label{sec/method/verifier}
Constraint verifiers are the realization of $f$, measuring how well the response satisfies the constraints. 
They take the model response (and prompt) as the input, returning a constraint satisfaction rate (CSR).
A higher CSR indicates that the response adheres to the constraints better.
The verifiers can be rule-based (e.g., a function comparing words) or model-based (e.g., a relevance scorer), typically easy to implement from scratch or adapt from existing tools.
In \Cref{sec/experiment}, we showcase the use of Python functions, model-based metrics, and rule engines as constraint verifiers.
Note that each task may be associated with one or more constraints. 
Thus, the complete constraint verifier could be a combination of multiple sub-verifiers. 
The final CSR will be a weighted average of CSR from each sub-verifier, with the weights determined by the importance of the constraints.

\subsection{Response Sampling}
\label{sec/method/response}
While a series of LM alignment studies have mentioned response sampling, little attention has been paid on improving the alignment effectiveness through decoding strageties. 
We draw inspiration from contrastive learning to gather high-quality negative responses \cite{robinson2021contrastive}. 
The key to this step is ensuring that responses for the same unlabeled instance are distinguishable by the constraint verifiers (i.e., true negative), while simultaneously achieving high sampling probability (i.e., hard negative).
If two responses have a close CSR, it could be challenging for even human annotators to decide which one is better.  
If the response with a low CSR also has a low sampling probability, penalizing it will not significantly benefit the model.
In a nutshell, we seek to collect high-probability responses with non-negligible CSR gaps.
Therefore, we employ decoding strategies that incorporate diversification and probability restriction, such as diverse beam search \cite{vijayakumar2018diverse}. %
This enables the collection of informative supervision signals in the next step.

\subsection{Constraint Verification}
\label{sec/method/verification}
Constraint verifiers can offer approximate but essential guidance for task adaptation, making them well-suited for the cost-efficient customization of LMs to specific tasks.
\MODEL takes advantage of this property of automatic constraint verifiers to provide supervision signals for LM alignment. %
Specifically, the constraint verifier returns a CSR for each response or response combination. %
Then, we can assign preference labels to responses for the same prompt based on their CSR.
For constraints defined over a single response or prompt-response pair, the response that has a higher CSR will be preferred. 
For example, in a task with label options constraint, a response within the option list is preferable to a response beyond it.
For constraints defined over multiple prompt-response pairs, \MODEL creates a response combination by picking one response for each prompt.
The constraint verifier computes the CSR for each response combination, and responses from the response combination with a higher CSR will be preferred. 
For example, when asking about events occurring before or after an event, the response combination that have no conflict (i.e., no overlap between the answers to `before' and `after') are preferable to those with conflicts. Then, each response will inherit the preference label of the combination it belongs to.
As a result, \MODEL can collect preference labels from constraint verifiers as supervision signals to align models based on any type of constraints introduced in \Cref{sec/method/constraint}.

\subsection{Training}
\label{sec/method/training}

With the preference labels from constraint verifiers as supervision signals, \MODEL follows the learning objective of \citet{yuan2023rrhf} with CSR as the reward. 
It encourages the model to generate the response with highest CSR for each prompt with
$$ \mathcal{L}_{ft} = - \sum_i \log P(y_i|\mathbf{x}, \mathbf{y}_{<i}),$$
and optimizes a rank loss over all responses for the same prompt based on their relative CSR
$$\mathcal{L}_{rank} = \sum_{CSR_i < CSR_j} \max (0, P(\mathbf{y}^i|x) - P(\mathbf{y}^j|x)).$$

Since the CSR gap between each response pair may indicate fine-grained preference information, such as the relevance score in text summarization, we can further enhance the above loss functions. 
For $\mathcal{L}_{ft}$, we use CSR to reweight each datapoint. 
Because the quality of the best responses we sample for different prompts may vary, this strategy amplifies the impact of responses with higher CSR while reducing noise.
For $\mathcal{L}_{rank}$, we use the CSR gap between each pair of responses as the ranking margin.
This strategy allows the ranking loss to consider the relative preference, providing more informative supervision signals. 

To further enhance learning efficiency, we adopt parameter-efficient tuning to align the LM with constraints.
Specifically, we train LoRA modules \cite{hu2021lora} as customized adapters in a plug-and-play manner.
The learning process is cost-efficient, and users have the flexibility to choose adapters based on constraints they need. 

%% file: section/3_experiment.tex
\section{Experiment}
\label{sec/experiment}

In this section, we evaluate \MODEL on three representative tasks, each of which has one distinct class of constraints introduced in \Cref{sec/method/constraint}, including fine-grained entity typing with label option and label hierarchy constraint ($f(y)$; \Cref{sec/experiment/fine}), abstractive summarization with document-summary relevance constraint ($f(x,y)$; \Cref{sec/experiment/abstractive}), and temporal question answering (QA) with the ``no temporal conflict'' constraint ($f(\{x_i,y_i\})$; \Cref{sec/experiment/temporal}). Moreover, we conduct a pilot study to verify the transferability of the constraint-following ability (\Cref{sec/experiment/constraint}).

\subsection{$f(y)$: Fine-Grained Entity Typing}
\label{sec/experiment/fine}

\stitle{Task and Constraint}
Fine-grained entity typing seeks to select one or more applicable entity types of different granularities for an entity in a given sentence. 
We select two sub-constraints defined over the model response for this task: 
(1) label option, requiring all entity types to be selected from a fixed option list; and
(2) label hierarchy, requiring to select a coarse-grained type if its corresponding fine-grained type is selected (e.g., an artist entity must be a person entity).
Verifying these constraints needs to check the model output $y$.
We implement the constraint verifier as a rule-based Python function, comparing the model response with the predefined label option and label hierarchy.
Its pseudo code is in \Cref{sec/appendix/verifier}.

\stitle{Dataset and Metric}
We conduct experiments on the FIGER dataset \cite{ling2012fine} consisting of 112 entity types in two granularities. %
We sample 1K instances, which is the smallest effective data size used for LM alignment in prior studies \cite{jin2023data,zhou2023lima}, from the official training set as the unlabeled data, and five additional instances as in-context examples. For evaluation, we use the official test set. Following \citet{ling2012fine}, we use macro-F1 over all instances as the evaluation metric. For this and the following tasks, we report the average result of three runs.

\begin{figure}[t]
    \centering
    \includegraphics[width=0.95\columnwidth]{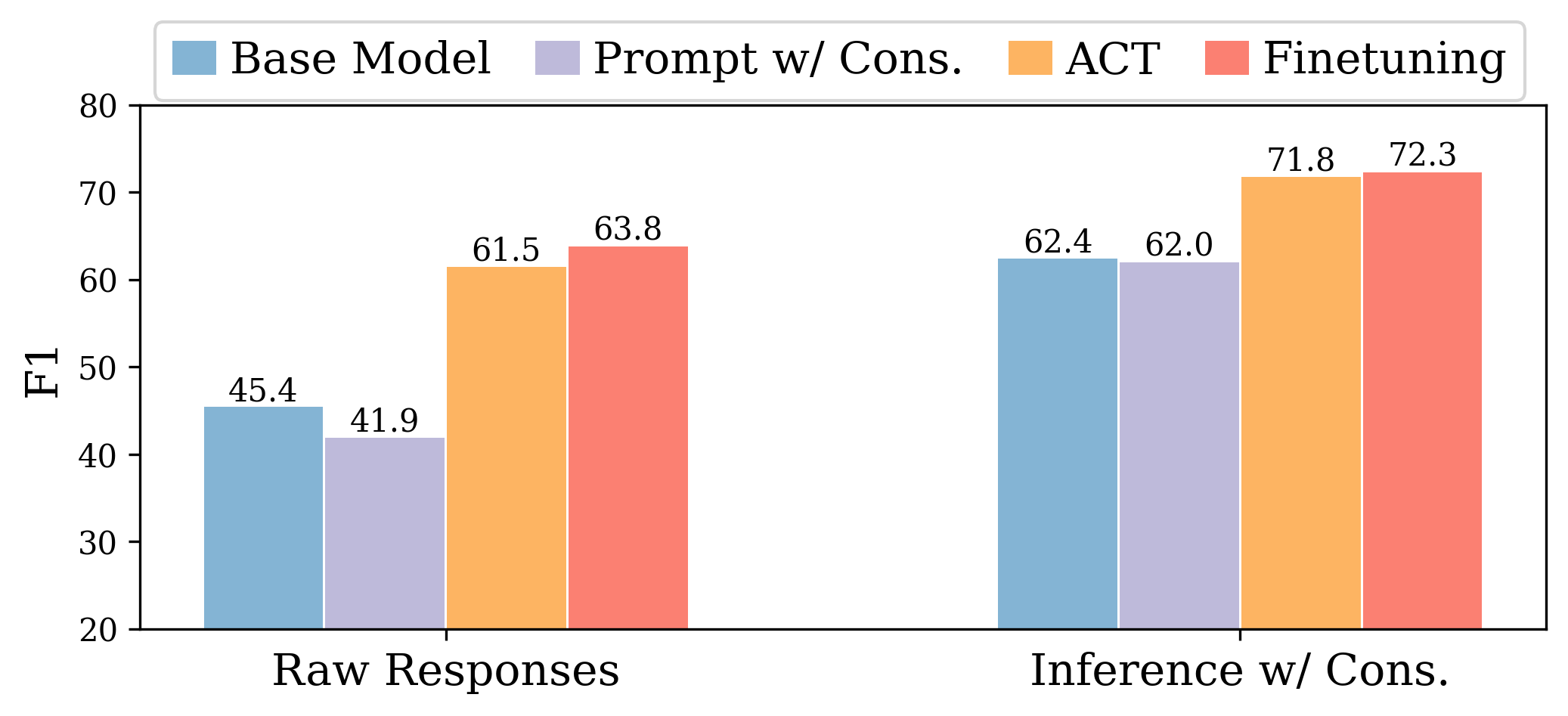}
    \caption{Results on fine-grained entity typing with $f(y)$ constraint. \MODEL, using supervision signals from automatic constraint verifiers, achieves performance close to that of  \textit{Finetuning} on the same amount of labeled data. 
    }
    \label{fig/figer}
    %\vspace{-1em}
\end{figure}

\stitle{Baselines}
We compare \MODEL with both training-free constraint integration and finetuning with labeled data. 
To integrate constraints into LMs, one way is \textit{prompt w/ constraints} by adding verbalized constraints in the prompt. It adds into prompts the list of entity types with \texttt{``Label options: \{all types\}"} and the type dependency with \texttt{``If an entity is any of \{fine-grained types\}, it must also be \{coarse-grained type\}."}
The other way is \textit{inference w/ constraints} through post-hoc correction.\footnote{While other inference-time constraint integration approaches may also work, we do not observe significant difference in performance.} The corrector is derived from the constraint verifier, correcting prediction errors according to the task constraints.
\textit{Finetuning} adopts the same instances used by \MODEL with human-annotated labels.

\begin{figure}[t]
    \centering
    \includegraphics[width=0.95\columnwidth]{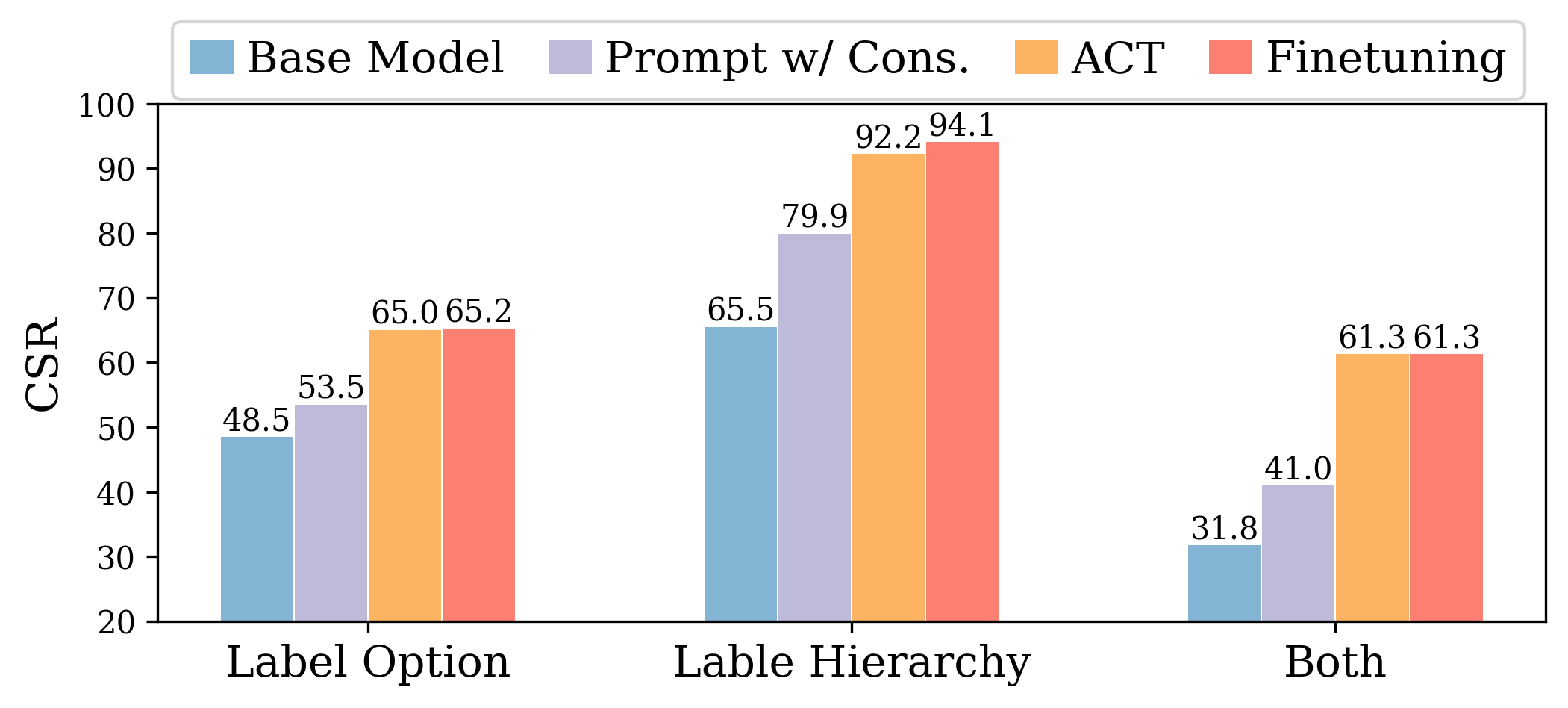}
    \caption{Average CSR of raw responses on fine-grained entity typing. \textit{Label Option} constraint limits the candidate set of entity types. \textit{Label Hierarchy} constraint requires the answer to follow the hierarchy between coarse- and fine-grained entity types. A correct answer must satisfy \textit{Both} constraints. \MODEL achieves CSR comparable to that of \textit{Finetuning}.}
    \label{fig/figer_csr}
    %\vspace{-1em}
\end{figure}

\stitle{Implementation Details}
For this and the following tasks, we use Falcon-7B-Instruct \cite{refinedweb} as the base model, because it is one of the few SOTA instruction-tuned LMs with Apache 2.0 license.
We apply LoRA tuning in both \MODEL and finetuning. 
All models are trained using the same prompt templates and hyper-parameters in \Cref{sec/appendix/prompt,sec/appendix/implementation}.
For each unlabeled instance, \MODEL collects multiple model responses through diverse beam search.
Note that in this task, we consider a binary CSR, selecting one response that satisfies all constraints and another that does not satisfies some constraints, for training.
During the training and inference for all methods, we use the same five in-context examples.

\stitle{Results}
As shown in \Cref{fig/figer}, \MODEL, with automatic feedback from constraint verifier, achieves comparable results to finetuning with human annotation on same amount of data.
Further analysis in \Cref{fig/figer_csr} shows that \MODEL achieves the same overall CSR as finetuning.
These observations indicate that feedback from automatic constraint verifiers are effective surrogate of human feedback.
Moreover, \MODEL can significantly improve the model's constraint-following capability with the help of automatic constraint verifiers.
Although inference w/ constraints can further improve the performance of all methods as a complement, the improvement on \MODEL and finetuning are much smaller, indicating most of the knowledge about label constraints are already learned during training. 
Prompt w/ constraints improves model CSR, but does not improve the F1 score. We attribute this to the increased prompt length. Verbalizing the constraint adds several hundreds of tokens in the prompt, which unsurprisingly make it more difficult to understand.

\input{table/human_evaluation}

\subsection{$f(x,y)$: Abstractive Summarization}
\label{sec/experiment/abstractive}

\stitle{Task and Constraint}
Abstractive summarization seeks to provide a brief summary for a given document.
An essential constraint for this task is relevance -- the information in the generated summary should be relevant to that in the given document. 
This constraint is necessary to achieve better factual consistency \cite{zhu-etal-2021-enhancing,dixit-etal-2023-improving} and information coverage. 
To verify this constraint, we need to compare the model input $x$ and output $y$.
We use BERTScore-Recall \cite{zhang2019bertscore} as the constraint verifier, because prior works have shown that it aligns well with the human judgement of summary quality and outperforms other metrics in downstream applications \cite{fabbri2021summeval,adlakha2023evaluating,gupta2023coverage}. Note that we compute the BERTScore-Recall between the model response and the input document as CSR, which allows \MODEL to collect feedback with no human-annotated summary.

\stitle{Dataset and Metrics}
We conduct experiments on the XSUM dataset \cite{narayan-etal-2018-dont}, where each news article is paired with a human-written one-sentence summary.
For training, we sample 1K instances from the official training set.
We evaluate the model performance in a zero-shot manner.
For automatic evaluation, we report ROUGE-L \cite{lin-2004-rouge}, BERTScore, and CSR. 
We further conduct \textbf{human evaluation} following the protocol in \citet{zhang2023benchmarking}. We recruit annotators from Amazon Mechanical Turk to label consistency (0 or 1), informativeness (5 point likert scale), and coherence (5 point likert scale) for system-generated and human-written summaries. Each summary is evaluated by three different annotators. The human evaluation instruction is in \Cref{sec/appendix/human_eval}.
Due to the computational and annotation cost, we sample 100 articles from the official test set for evaluation. 

\stitle{Baselines}
\textit{Prompt w/ constraints} emphasizes that \texttt{``Your summary should be relevant to the input document"} in the prompt.
\textit{Inference w/ constraints} adopts the constraint verifier to rerank multiple sampled summaries, which is shown to outperform some training-based methods in prior work \cite{cao-wang-2021-cliff}. 
\textit{Finetuning} trains the LM with human-written summaries on the same training instances as \MODEL.
Note that inference w/ constraints is complementary to other approaches, so we also apply it to \MODEL and finetuning.

\stitle{Implementation Details}
For \MODEL, we have two variants, with and without model warmup on 100 human-labeled data.
With only a small amount of labeled data, the warm-up step enables the model to generate reasonable responses for a relatively complicated task, even though the model still achieves relatively low performance.
We use the enhanced loss function, where $l_{ft}$ is re-weighted and $l_{rank}$ has a ranking margin.
More details are in \Cref{sec/appendix/implementation}.

\begin{figure}[t]
    \centering
    \includegraphics[width=0.95\columnwidth]{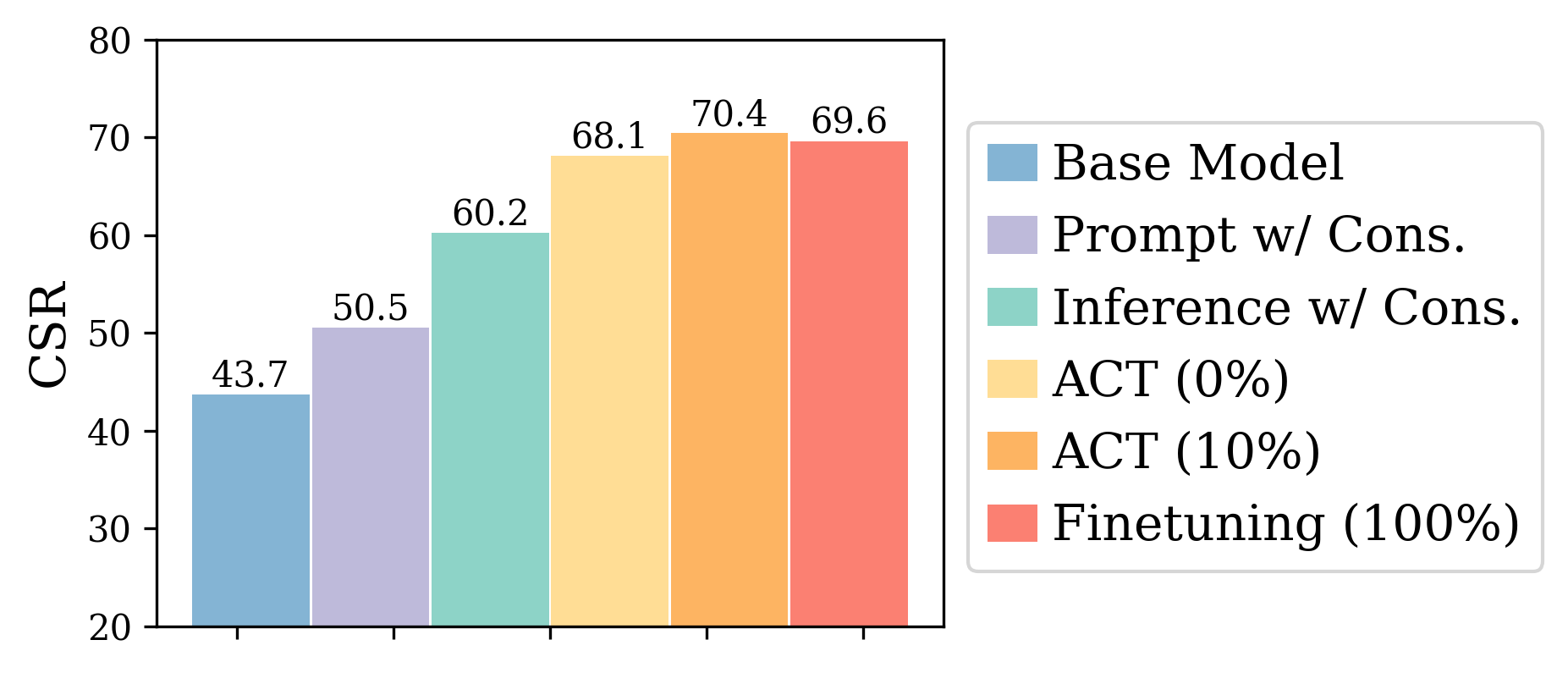}
    \caption{Average CSR of relevance constraint on model-generated summaries. \MODEL achieves even higher CSR than \textit{Finetuning}.}
    \label{fig/xsum}
    %\vspace{-1em}
\end{figure}

\stitle{Results}
As shown in \Cref{table/xsum}, \MODEL with model warmup achieves comparable results in comparison with finetuning, and even outperforms the latter in terms of BERTScore in automatic evaluation and coherence in human evaluation. \MODEL with no human-labeled data, also performs as well as finetuning in terms of factual consistency. Both human and automatic evaluation indicate that aligning the model with the automatically verifiable relevance constraint can enhance the model performance on text summarization. Although model-generated summaries still have a gap with ground-truth summaries, it will not be difficult to scale up the size of training data for \MODEL with the help of the automatic constraint verifier.
We further analyze model CSR in \Cref{fig/xsum}. \MODEL with warmup also outperforms finetuning from the perspective of constraint satisfaction. Both \MODEL and finetuning significantly outperforms the base model. 
This observation indicates a positive correlation between the quality of summaries and the adherence level to the summary-document relevance constraint.

\subsection{$f(\{x_i,y_i\})$: Temporal QA}
\label{sec/experiment/temporal}

\stitle{Task and Constraint}
Temporal question answering seeks to answer questions about the temporal relationship of events based on a given passage. 
Due to the nature of time, the responses to several interconnected questions should not have temporal conflicts. For example, the answers to \textit{"what happens before event A"} and \textit{"what happens after event A"} should have no overlap. Otherwise, an event may occur both before and after event A, leading to a time cycle. 
This constraint requires to compare multiple question-answer pairs $\{x_i, y_i\}$.
We define a rule engine in Python as the constraint verifier, which identifies conflicts in temporal relationships among events.

\stitle{Dataset and Metrics}
We conduct experiments on the TORQUE dataset \cite{ning-etal-2020-torque}, where each passage is paired with multiple temporal questions. We focus on the default set of questions which have clear logical relationships asking what happens before/during/after an event according to a given passage. We sample 1K group of questions from the official training set, leading to 3K instances in total.
We report the average macro- and micro-F1 of three runs on the official development set.

\stitle{Baselines}
Due to the complexity of the task and constraint, the raw model cannot generate reasonable responses and simply integrating constraints into the prompt or the inference process does not make the situation better. Therefore, we mainly compare our method with \textit{finetuning} on human-annotated QA pairs.

\stitle{Implementation Details}
Since the base model fails to give reasonable answers, we apply model warmup for all methods. Specifically, we use 1K labeled data to warmup the model before \MODEL or further finetuning.\footnote{
This is similar to the SFT-RLHF paradigm \cite{ouyang2022training}, where the relatively light-weight SFT step warms up the model to provide reasonable responses for data gathering to support RLHF.
The warmup step helps to mitigate the ``garbage in, garbage out" problem, ensuring the availability of relatively good responses to facilitate informative feedback, particularly for complex tasks.}
Then, \MODEL further tunes the model on 1K unlabeled data with feedback from the constraint verifier, while further finetuning adopts additional 1K human-labeled data.
When collecting feedback from the constraint verifier, we sample 2 responses for each instance. Then, for all the $2^k$ response combinations of $k$ related questions, we use the constraint verifier to find one with no or the least conflicts as the preferred response combination. We use the preference label of the response combination as the preference label of each response within this combination.
For all methods, we use the same three in-context examples.
More details are in \Cref{sec/appendix/prompt,sec/appendix/implementation}.

\begin{figure}
  \centering
  \includegraphics[width=0.49\columnwidth]{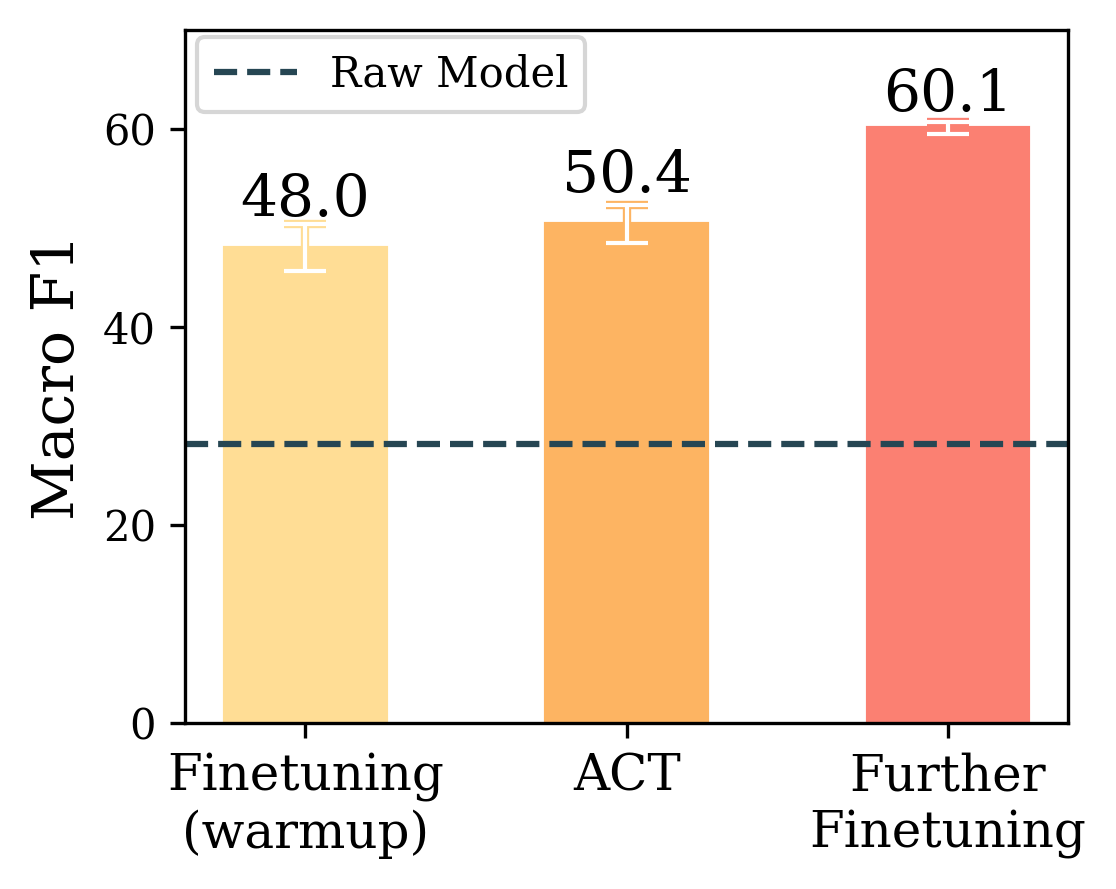}
  \includegraphics[width=0.49\columnwidth]{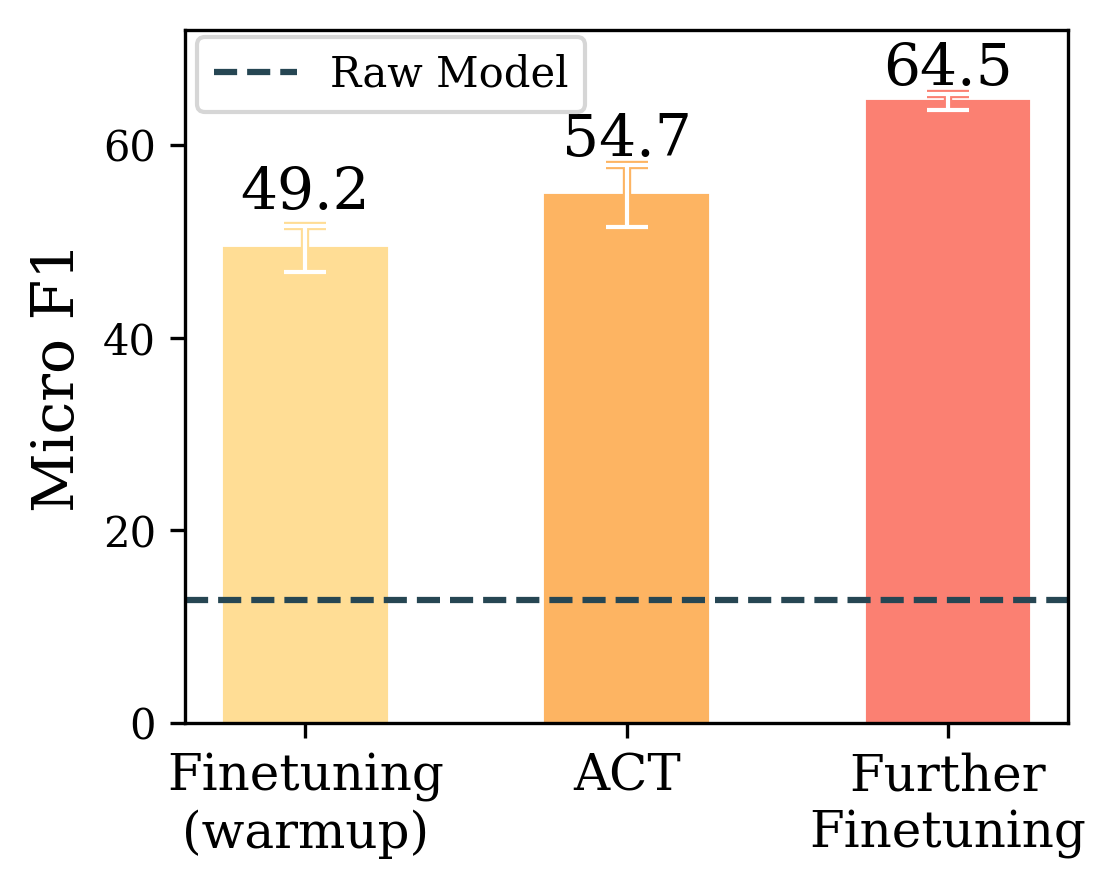}
  \caption{Results on temporal QA with constraint of $f(\{x_i, y_i\})$ class. As the raw model cannot generate reasonable answers, we use \textit{Finetuning (warmup)} as the base model. \MODEL can even improve the performance of a finetuned model. \textit{Further Finetuning} continually train the base model on labeled data.\footnotemark %
  }
  \label{fig/torque}
  %\vspace{-1em}
\end{figure}

\footnotetext{
    Since this experiment seeks to evaluate \MODEL on a specific class of constraints, %
    we do not consider other stronger constraints. 
    The ``no temporal conflict'' constraint only provides weak approximation of the answers. %
    Thus, not surprisingly, further finetuning achieves better performance.}

\stitle{Results}
As shown in \Cref{fig/torque}, the base model totally fails to give reasonable responses, revealing the difficulty of the task.
\MODEL improves the performance of the warmed-up model by 2.4 points in terms of macro-F1 and 5.5 points in terms of micro-F1. 
This indicates that \MODEL can even improve the performance of a finetuned model.

\subsection{Constraint Transfer}
\label{sec/experiment/constraint}

To verify the transferability of constraint-following capability, we apply \MODEL to train and test the LM on different tasks with the same type of constraint.

\stitle{Task and Constraint}
We conduct experiments on the extractiveness constraint, where the model response must be extracted from the input.
We select three tasks with this constraint: entity extraction, slot extraction, and event trigger extraction.
The pseudo code of constraint verifier is in \Cref{sec/appendix/verifier}

\stitle{Dataset and Metrics}
We use FIGER for entity extraction, MASSIVE \cite{fitzgerald2022massive} for slot extraction, and ACE 2005 \cite{walker2006ace} for event trigger extraction. 
We sample 1K instances from each of MASSIVE and FIGER for training and 2K instances from ACE 2005 for evaluation.
The CSR shows the model capability of following the target constraint. 

\stitle{Implementation Details}
Prompts for all tasks adopt the same format with a constraint ``You must extract the answer from the input sentence.''
During training and inference, we use five additional in-context examples.
Detailed prompts and hyper-parameters can be found in \Cref{sec/appendix/prompt,sec/appendix/implementation}.

\input{table/transfer}

\stitle{Results}
Results in \Cref{table/transfer} show that the extractiveness constraint learned from entity extraction and slot extraction can be transferred to event trigger extraction, resulting in an improvement in CSR ranging from 8.9\% to 17.4\%, respectively.
This indicates that the constraint-following capability is transferable.
Combining multiple source tasks leads to better performance.

%% file: table/human_evaluation.tex
\begin{table*}
  \centering
  \small
  \setlength{\tabcolsep}{5pt}
  \begin{tabular}{cc|cc|ccc}
    \toprule
     \multirow{2}{*}{\textbf{Method}} & \textbf{Training Data} & \multicolumn{2}{c}{\underline{\textbf{Automic Evaluation}}} & \multicolumn{3}{c}{\underline{\textbf{Human Evaluation}}} \\
     & labeled~:~unlabeled & \textbf{BERTScore} & \textbf{ROUGE-L} & \textbf{Consistency} & \textbf{Informativeness} & \textbf{Coherence}   \\ %
    \midrule
    Raw Model & -  & 42.8 & 10.7 & 0.54 & 2.78 & 2.93  \\
    Prompt w/ Cons. & - & 55.5 & 12.8 &  0.63 & 3.06 & 3.21   \\
    Inference w/ Cons. & - & 58.9 & 13.6 & 0.56 & 2.87 & 3.07 \\
    \MODEL & \textcolor{red}{0\%}~:~\textcolor{blue}{100\%} & 65.1 & 15.7 &  \textbf{0.68} & 3.12 & 3.35   \\
    \MODEL & \textcolor{red}{10\%}~:~\textcolor{blue}{90\%} & \textbf{68.6} & \textbf{18.2} &  0.65 & 3.20 & \textbf{3.44}  \\
    Finetuning & \textcolor{red}{100\%}~:~\textcolor{blue}{0\%} & 68.2 & \textbf{18.2} &  \textbf{0.68} & \textbf{3.24} & 3.40  \\ \midrule
    Ground-Truth & - & - & - & 0.81 & 3.66 & 3.81 \\
    \bottomrule
  \end{tabular}
  \caption{
    Automatic and human evaluation on abstractive summarization with constraint of $f(x,y)$ class. We also report the ratio of human-labeled and unlabeled training data for \MODEL and \textit{Finetuning}. Note that \textit{Inference w/ Constraints} is also applied to \MODEL and \textit{Finetuning}, as they are complementary.
  }
  \label{table/xsum}
  % \vspace{-1.5em}
\end{table*}

%% file: table/transfer.tex
\begin{table}[t]
  \centering
  \small
  \begin{tabular}{lc} %
    \toprule
    \textbf{Source Task} & \textbf{CSR on Target Task (T3)} \\
    \midrule
    - & 58.8 \\ \midrule
    Slot Extraction (T1) & 67.7 \\
    Entity Extraction (T2) & 73.9 \\
    Both (T1+T2) & 76.2 \\
    \bottomrule
  \end{tabular}
  \caption{CSR of extractiveness constraint on event trigger extraction (T3). Learning the constraint from other tasks (T1 \& T2) can improve the CSR on the target task.}
  \label{table/transfer}
  % \vspace{-1em}
\end{table}

%% file: section/5_related_work.tex
\section{Related Work}
We briefly review two relevant research topics.

\stitle{Constraints in NLP}
Constraints provide essential information about the detailed requirements of user intents, which widely exist in various NLP tasks, such as natural language inference \cite{roth2004linear,minervini-riedel-2018-adversarially,li-etal-2019-logic}, information extraction \cite{ning-etal-2017-structured,wang-etal-2020-joint,lin-etal-2023-global}, and text summarization \cite{dou-etal-2021-gsum,wang-etal-2022-salience,dixit-etal-2023-improving}. Constraints in these tasks range from simple fixed label options and format requirements to complex logic dependency \cite{faghihi2023gluecons}.
Prior works have integrated these constraints into artificial intelligent models through learning-based or inference-only methods, such as constraint driven learning \cite{chang-etal-2007-guiding,minervini-riedel-2018-adversarially}, structured inference \cite{ning-etal-2017-structured,wang2023regularization}, and constrained decoding \cite{hokamp-liu-2017-lexically,qin2022cold}. 
Recent work also investigated integrating constraints into LMs to improve model performance on binary question answering \cite{burns2022discovering,jung-etal-2022-maieutic} and natural language inference \cite{mitchell-etal-2022-enhancing}. 
Building upon these findings, we utilize constraints indicated in task instructions to align LMs to various user intents with an unified and cost-efficient framework.

\stitle{Language Model Alignment}
The success of LMs has brought considerable attention to language model alignment recently \cite{ouyang2022training}. 
In terms of alignment data, early works in this direction primarily centered around aligning to human feedback, including human-annotated instruction-response pairs and human preference of model responses \cite{ouyang2022training,bai2022training,wang-etal-2022-super,longpre2023flan}. More recent research has delved into self-generated feedback from LMs \cite{wang-etal-2023-self-instruct,li2023self}. We investigate a novel source of automatic feedback, specifically for LM adaptation. %
In terms of training method, prior works have aligned LMs to human intents through reinforcement learning \cite{ouyang2022training} or rank-based learning \cite{yuan2023rrhf,rafailov2023direct,liu2023chain}. Our learning objective is derived from rank-based learning, since its training process is more stable.

%% file: section/6_conclusion.tex
\section{Conclusion}

In this paper, we propose an efficient and unified LM alignment framework, \MODEL, to adapt LMs to customized tasks with various constraints. \MODEL relies on constraint verifiers, which are typically easy to implement, to automatically provide CSR as supervision signals. \MODEL can effectively enhance LMs' capability to adhere to constraints in user instructions, thereby fulfilling user intents. We investigate common constraints in NLP tasks, categorize them into three classes based on the types of their arguments, and verify the effectiveness of \MODEL on all classes of constraints. 
Experiments on constraint transfer further shows the feasibility of tuning general constraint-following LMs.

\section*{Limitations}
Due to license and accessibility constraints, we cannot verify the effectiveness of \MODEL across a wide range of LMs. Despite the similarities in model structures and training processes among these LMs, variations in their implementation details may result in slightly different performance gains when applying \MODEL. 
Furthermore, while \MODEL notably reduces the cost of data collection for custom tasks, the steps involving constraint selection and verifier realization still require human effort. Automating these steps would contribute to further improvements.
Finally, while our work demonstrates the potential of training various constraint-following adapters and general constraint-following models, we acknowledge that there is ample room for further exploration in this expansive area, providing opportunities for future research.

%% file: section/7_appendix.tex
\clearpage

\appendix

\section{Constraint Verifiers}
\label{sec/appendix/verifier}
We present the constraint verifiers in pseudo code of Python style.

\stitle{Label Option and Hierarchy}
\begin{tcolorbox}
\small
\begin{minted}{python}
# OPTIONS is a fixed list of valid options
# FINE2COARSE is a map from each 
# fine-grained entity type to its 
# corresponding coarse-grained entity type

def label_option(answers):
    for x in answers:
        if x not in OPTIONS:
            return 0
    return 1

def label_hierarchy(answers):
    for x in answers:
        if x not in FINE2COARSE: 
            continue
        if FINE2COARSE[x] not in answers:
            return 0
    return 1

def constraint_verifier(response):
    answers = response.split(", ")
    first_cons = label_option(answers)
    second_cons = label_hierarchy(answers)
    return min(first_cons, second_cons)
\end{minted}
\end{tcolorbox}

\stitle{Extractiveness}
\begin{tcolorbox}
\small
\begin{minted}{python}
def constraint_verifier(inputx, response):
    csr = int(response in inputx)
    return csr
\end{minted}
\end{tcolorbox}

\section{Prompt Template}
\label{sec/appendix/prompt}

We follow the prompt template of \citet{taori2023alpaca} for all experiments:

\begin{tcolorbox}[colback=white, colframe=black!75!black, boxrule=0.5pt, sharp corners, title=TEMPLATE]
Below is an instruction that describes a task. Write a response that appropriately completes the request. \\
\#\#\# Instruction: \\
\texttt{\{\$INSTRUCTION\}} \\ \\
\#\#\# Input: \\
\texttt{\{\$INPUT\}} \\ \\
\#\#\# Response: \\
\texttt{\{\$RESPONSE\}}
\end{tcolorbox}

\stitle{Fine-Grained Entity Typing}
\begin{tcolorbox}[colback=white, colframe=black!75!black, boxrule=0.5pt, sharp corners, title=INSTRUCTION]
List all entity types of an entity in a given sentence. \\
Options: \texttt{\{\$OPTIONS\}}. \\
If the entity is any of \texttt{\{\$FINETYPES\}}, it is also \texttt{\{\$COARSETYPE\}}.
\end{tcolorbox}
\begin{tcolorbox}[colback=white, colframe=black!75!black, boxrule=0.5pt, sharp corners, title=INPUT]
In the sentence \texttt{\{\$SENTENCE\}}, what are the types of the entity \texttt{\{\$ENTITY\}}?
\end{tcolorbox}

\stitle{Abstractive Summarization}
\begin{tcolorbox}[colback=white, colframe=black!75!black, boxrule=0.5pt, sharp corners, title=INSTRUCTION]
Please generate a one-sentence summary for the given document.
\end{tcolorbox}
\begin{tcolorbox}[colback=white, colframe=black!75!black, boxrule=0.5pt, sharp corners, title=INPUT]
\texttt{\{\$DOCUMENT\}}
\end{tcolorbox}

\stitle{Temporal QA}
\begin{tcolorbox}[colback=white, colframe=black!75!black, boxrule=0.5pt, sharp corners, title=INSTRUCTION]
Select the best options to answer the question according to the passage.
\end{tcolorbox}
\begin{tcolorbox}[colback=white, colframe=black!75!black, boxrule=0.5pt, sharp corners, title=INPUT]
Passage: \texttt{\{\$PASSAGE\}} \\
Question: \texttt{\{\$QUESTION\}} \\
Options: \texttt{\{\$OPTIONS\}}
\end{tcolorbox}

\stitle{Constraint Transfer}
\begin{tcolorbox}[colback=white, colframe=black!75!black, boxrule=0.5pt, sharp corners, title=INSTRUCTION]
Identify the [entity / slot / event trigger] in the given sentence. \\
Your response must directly indicate the target information. \\
You must extract the answer from the input sentence.
\end{tcolorbox}
\begin{tcolorbox}[colback=white, colframe=black!75!black, boxrule=0.5pt, sharp corners, title=INPUT]
Which words indicate \texttt{\{\$TYPE\}} in the sentence \texttt{\{\$SENTENCE\}}.
\end{tcolorbox}

\section{Hyper-parameters}
\label{sec/appendix/implementation}
We use the same hyperparameters in all experiments unless otherwise specified.

\stitle{Training}
We train the models for 10 epochs with a batch size of 32 and a constant learning rate of 1e-5. 
We apply LoRA modules to the query, key, and value projectors in the attention module of each Transformer layer.
The LoRA alpha, LoRA rank, and LoRA dropout are set to 16, 64, and 0.1 respectively.
Following \citet{yuan2023rrhf}, we do not adjust the coefficient between $L_{ft}$ and $L_{rank}$, but simply add them.
All inputs are left padded to 1,024 tokens. 
Note that we sampled 10\% of the collected data for validation.
For constraint transfer, we enlarge the size of LoRA modules and the learning rate to accommodate the shared constraint knowledge from different tasks.
Specifically, we set LoRA alpha to 32, LoRA rank to 64, and constant learning rate to 2e-5.

\stitle{Inference}
During evaluation, we apply greedy decoding.
For response sampling, we apply diverse beam search with four beams, four beam groups, and a diversity penalty of 1.

\section{Human Evaluation}
\label{sec/appendix/human_eval}

\begin{figure*}[h]
    \centering
    \includegraphics[width=\textwidth]{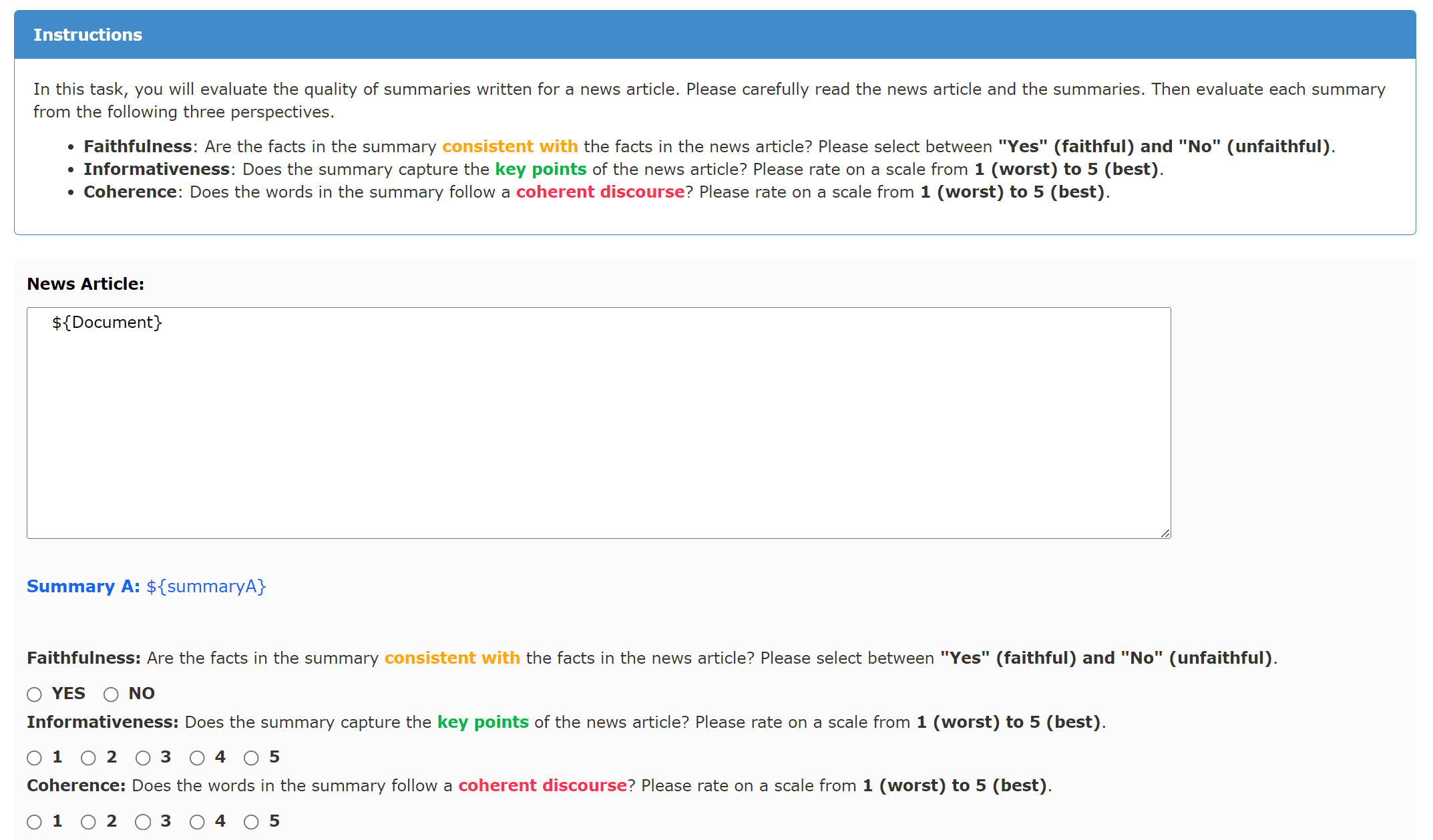}
    \caption{Human evaluation interface.}
    \label{fig/human_eval}
\end{figure*}

The interface including instructions for human evaluation is shown in \Cref{fig/human_eval}.

%% file: main.bbl
\begin{thebibliography}{54}
\expandafter\ifx\csname natexlab\endcsname\relax\def\natexlab#1{#1}\fi

\bibitem[{Abdin et~al.(2023)Abdin, Gunasekar, Chandrasekaran, Li, Yuksekgonul, Peshawaria, Naik, and Nushi}]{abdin2023kitab}
Marah~I Abdin, Suriya Gunasekar, Varun Chandrasekaran, Jerry Li, Mert Yuksekgonul, Rahee~Ghosh Peshawaria, Ranjita Naik, and Besmira Nushi. 2023.
\newblock Kitab: Evaluating llms on constraint satisfaction for information retrieval.
\newblock \emph{arXiv preprint arXiv:2310.15511}.

\bibitem[{Adlakha et~al.(2023)Adlakha, BehnamGhader, Lu, Meade, and Reddy}]{adlakha2023evaluating}
Vaibhav Adlakha, Parishad BehnamGhader, Xing~Han Lu, Nicholas Meade, and Siva Reddy. 2023.
\newblock Evaluating correctness and faithfulness of instruction-following models for question answering.
\newblock \emph{arXiv preprint arXiv:2307.16877}.

\bibitem[{Bai et~al.(2022)Bai, Jones, Ndousse, Askell, Chen, DasSarma, Drain, Fort, Ganguli, Henighan et~al.}]{bai2022training}
Yuntao Bai, Andy Jones, Kamal Ndousse, Amanda Askell, Anna Chen, Nova DasSarma, Dawn Drain, Stanislav Fort, Deep Ganguli, Tom Henighan, et~al. 2022.
\newblock Training a helpful and harmless assistant with reinforcement learning from human feedback.
\newblock \emph{arXiv preprint arXiv:2204.05862}.

\bibitem[{Burns et~al.(2022)Burns, Ye, Klein, and Steinhardt}]{burns2022discovering}
Collin Burns, Haotian Ye, Dan Klein, and Jacob Steinhardt. 2022.
\newblock Discovering latent knowledge in language models without supervision.
\newblock In \emph{The Eleventh International Conference on Learning Representations}.

\bibitem[{Cao and Wang(2021)}]{cao-wang-2021-cliff}
Shuyang Cao and Lu~Wang. 2021.
\newblock \href {https://doi.org/10.18653/v1/2021.emnlp-main.532} {{CLIFF}: Contrastive learning for improving faithfulness and factuality in abstractive summarization}.
\newblock In \emph{Proceedings of the 2021 Conference on Empirical Methods in Natural Language Processing}, pages 6633--6649, Online and Punta Cana, Dominican Republic. Association for Computational Linguistics.

\bibitem[{Chang et~al.(2007)Chang, Ratinov, and Roth}]{chang-etal-2007-guiding}
Ming-Wei Chang, Lev Ratinov, and Dan Roth. 2007.
\newblock \href {https://aclanthology.org/P07-1036} {Guiding semi-supervision with constraint-driven learning}.
\newblock In \emph{Proceedings of the 45th Annual Meeting of the Association of Computational Linguistics}, pages 280--287, Prague, Czech Republic. Association for Computational Linguistics.

\bibitem[{Dixit et~al.(2023)Dixit, Wang, and Chen}]{dixit-etal-2023-improving}
Tanay Dixit, Fei Wang, and Muhao Chen. 2023.
\newblock \href {https://doi.org/10.18653/v1/2023.acl-short.78} {Improving factuality of abstractive summarization without sacrificing summary quality}.
\newblock In \emph{Proceedings of the 61st Annual Meeting of the Association for Computational Linguistics (Volume 2: Short Papers)}, pages 902--913, Toronto, Canada. Association for Computational Linguistics.

\bibitem[{Dou et~al.(2021)Dou, Liu, Hayashi, Jiang, and Neubig}]{dou-etal-2021-gsum}
Zi-Yi Dou, Pengfei Liu, Hiroaki Hayashi, Zhengbao Jiang, and Graham Neubig. 2021.
\newblock \href {https://doi.org/10.18653/v1/2021.naacl-main.384} {{GS}um: A general framework for guided neural abstractive summarization}.
\newblock In \emph{Proceedings of the 2021 Conference of the North American Chapter of the Association for Computational Linguistics: Human Language Technologies}, pages 4830--4842, Online. Association for Computational Linguistics.

\bibitem[{Fabbri et~al.(2021)Fabbri, Kry{\'s}ci{\'n}ski, McCann, Xiong, Socher, and Radev}]{fabbri2021summeval}
Alexander~R Fabbri, Wojciech Kry{\'s}ci{\'n}ski, Bryan McCann, Caiming Xiong, Richard Socher, and Dragomir Radev. 2021.
\newblock Summeval: Re-evaluating summarization evaluation.
\newblock \emph{Transactions of the Association for Computational Linguistics}, 9:391--409.

\bibitem[{Faghihi et~al.(2023)Faghihi, Nafar, Zheng, Mirzaee, Zhang, Uszok, Wan, Premsri, Roth, and Kordjamshidi}]{faghihi2023gluecons}
Hossein~Rajaby Faghihi, Aliakbar Nafar, Chen Zheng, Roshanak Mirzaee, Yue Zhang, Andrzej Uszok, Alexander Wan, Tanawan Premsri, Dan Roth, and Parisa Kordjamshidi. 2023.
\newblock Gluecons: A generic benchmark for learning under constraints.
\newblock \emph{AAAI 2023}.

\bibitem[{FitzGerald et~al.(2022)FitzGerald, Hench, Peris, Mackie, Rottmann, Sanchez, Nash, Urbach, Kakarala, Singh et~al.}]{fitzgerald2022massive}
Jack FitzGerald, Christopher Hench, Charith Peris, Scott Mackie, Kay Rottmann, Ana Sanchez, Aaron Nash, Liam Urbach, Vishesh Kakarala, Richa Singh, et~al. 2022.
\newblock Massive: A 1m-example multilingual natural language understanding dataset with 51 typologically-diverse languages.
\newblock \emph{arXiv preprint arXiv:2204.08582}.

\bibitem[{Gupta et~al.(2023)Gupta, Singh, and Gardner}]{gupta2023coverage}
Shivanshu Gupta, Sameer Singh, and Matt Gardner. 2023.
\newblock Coverage-based example selection for in-context learning.
\newblock \emph{arXiv preprint arXiv:2305.14907}.

\bibitem[{Hokamp and Liu(2017)}]{hokamp-liu-2017-lexically}
Chris Hokamp and Qun Liu. 2017.
\newblock \href {https://doi.org/10.18653/v1/P17-1141} {Lexically constrained decoding for sequence generation using grid beam search}.
\newblock In \emph{Proceedings of the 55th Annual Meeting of the Association for Computational Linguistics (Volume 1: Long Papers)}, pages 1535--1546, Vancouver, Canada. Association for Computational Linguistics.

\bibitem[{Hu et~al.(2021)Hu, Wallis, Allen-Zhu, Li, Wang, Wang, Chen et~al.}]{hu2021lora}
Edward~J Hu, Phillip Wallis, Zeyuan Allen-Zhu, Yuanzhi Li, Shean Wang, Lu~Wang, Weizhu Chen, et~al. 2021.
\newblock Lora: Low-rank adaptation of large language models.
\newblock In \emph{International Conference on Learning Representations}.

\bibitem[{Jiang et~al.(2023)Jiang, Wang, Zeng, Zhong, Li, Mi, Shang, Jiang, Liu, and Wang}]{jiang2023followbench}
Yuxin Jiang, Yufei Wang, Xingshan Zeng, Wanjun Zhong, Liangyou Li, Fei Mi, Lifeng Shang, Xin Jiang, Qun Liu, and Wei Wang. 2023.
\newblock Followbench: A multi-level fine-grained constraints following benchmark for large language models.
\newblock \emph{arXiv preprint arXiv:2310.20410}.

\bibitem[{Jin et~al.(2023)Jin, Mehri, Hazarika, Padmakumar, LEE, Liu, and Namazifar}]{jin2023data}
Di~Jin, Shikib Mehri, Devamanyu Hazarika, Aishwarya Padmakumar, SUNGJIN LEE, Yang Liu, and Mahdi Namazifar. 2023.
\newblock Data-efficient alignment of large language models with human feedback through natural language.
\newblock In \emph{NeurIPS 2023 Workshop on Instruction Tuning and Instruction Following}.

\bibitem[{Jung et~al.(2022)Jung, Qin, Welleck, Brahman, Bhagavatula, Le~Bras, and Choi}]{jung-etal-2022-maieutic}
Jaehun Jung, Lianhui Qin, Sean Welleck, Faeze Brahman, Chandra Bhagavatula, Ronan Le~Bras, and Yejin Choi. 2022.
\newblock \href {https://doi.org/10.18653/v1/2022.emnlp-main.82} {Maieutic prompting: Logically consistent reasoning with recursive explanations}.
\newblock In \emph{Proceedings of the 2022 Conference on Empirical Methods in Natural Language Processing}, pages 1266--1279, Abu Dhabi, United Arab Emirates. Association for Computational Linguistics.

\bibitem[{Li et~al.(2019)Li, Gupta, Mehta, and Srikumar}]{li-etal-2019-logic}
Tao Li, Vivek Gupta, Maitrey Mehta, and Vivek Srikumar. 2019.
\newblock \href {https://doi.org/10.18653/v1/D19-1405} {A logic-driven framework for consistency of neural models}.
\newblock In \emph{Proceedings of the 2019 Conference on Empirical Methods in Natural Language Processing and the 9th International Joint Conference on Natural Language Processing (EMNLP-IJCNLP)}, pages 3924--3935, Hong Kong, China. Association for Computational Linguistics.

\bibitem[{Li et~al.(2023)Li, Yu, Zhou, Schick, Zettlemoyer, Levy, Weston, and Lewis}]{li2023self}
Xian Li, Ping Yu, Chunting Zhou, Timo Schick, Luke Zettlemoyer, Omer Levy, Jason Weston, and Mike Lewis. 2023.
\newblock Self-alignment with instruction backtranslation.
\newblock \emph{arXiv preprint arXiv:2308.06259}.

\bibitem[{Lin(2004)}]{lin-2004-rouge}
Chin-Yew Lin. 2004.
\newblock \href {https://aclanthology.org/W04-1013} {{ROUGE}: A package for automatic evaluation of summaries}.
\newblock In \emph{Text Summarization Branches Out}, pages 74--81, Barcelona, Spain. Association for Computational Linguistics.

\bibitem[{Lin et~al.(2023)Lin, Zhang, and Song}]{lin-etal-2023-global}
Zizheng Lin, Hongming Zhang, and Yangqiu Song. 2023.
\newblock \href {https://doi.org/10.18653/v1/2023.findings-eacl.191} {Global constraints with prompting for zero-shot event argument classification}.
\newblock In \emph{Findings of the Association for Computational Linguistics: EACL 2023}, pages 2527--2538, Dubrovnik, Croatia. Association for Computational Linguistics.

\bibitem[{Ling and Weld(2012)}]{ling2012fine}
Xiao Ling and Daniel Weld. 2012.
\newblock Fine-grained entity recognition.
\newblock In \emph{Proceedings of the AAAI Conference on Artificial Intelligence}, volume~26, pages 94--100.

\bibitem[{Liu et~al.(2023)Liu, Sferrazza, and Abbeel}]{liu2023chain}
Hao Liu, Carmelo Sferrazza, and Pieter Abbeel. 2023.
\newblock Chain of hindsight aligns language models with feedback.
\newblock \emph{arXiv preprint arXiv:2302.02676}, 3.

\bibitem[{Liu et~al.(2022)Liu, Liu, Radev, and Neubig}]{liu-etal-2022-brio}
Yixin Liu, Pengfei Liu, Dragomir Radev, and Graham Neubig. 2022.
\newblock \href {https://doi.org/10.18653/v1/2022.acl-long.207} {{BRIO}: Bringing order to abstractive summarization}.
\newblock In \emph{Proceedings of the 60th Annual Meeting of the Association for Computational Linguistics (Volume 1: Long Papers)}, pages 2890--2903, Dublin, Ireland. Association for Computational Linguistics.

\bibitem[{Longpre et~al.(2023)Longpre, Hou, Vu, Webson, Chung, Tay, Zhou, Le, Zoph, Wei et~al.}]{longpre2023flan}
Shayne Longpre, Le~Hou, Tu~Vu, Albert Webson, Hyung~Won Chung, Yi~Tay, Denny Zhou, Quoc~V Le, Barret Zoph, Jason Wei, et~al. 2023.
\newblock The flan collection: Designing data and methods for effective instruction tuning.
\newblock In \emph{Proceedings of the 40 th International Conference on Machine Learning}.

\bibitem[{Minervini and Riedel(2018)}]{minervini-riedel-2018-adversarially}
Pasquale Minervini and Sebastian Riedel. 2018.
\newblock \href {https://doi.org/10.18653/v1/K18-1007} {Adversarially regularising neural {NLI} models to integrate logical background knowledge}.
\newblock In \emph{Proceedings of the 22nd Conference on Computational Natural Language Learning}, pages 65--74, Brussels, Belgium. Association for Computational Linguistics.

\bibitem[{Mishra et~al.(2022)Mishra, Khashabi, Baral, and Hajishirzi}]{mishra-etal-2022-cross}
Swaroop Mishra, Daniel Khashabi, Chitta Baral, and Hannaneh Hajishirzi. 2022.
\newblock \href {https://doi.org/10.18653/v1/2022.acl-long.244} {Cross-task generalization via natural language crowdsourcing instructions}.
\newblock In \emph{Proceedings of the 60th Annual Meeting of the Association for Computational Linguistics (Volume 1: Long Papers)}, pages 3470--3487, Dublin, Ireland. Association for Computational Linguistics.

\bibitem[{Mitchell et~al.(2022)Mitchell, Noh, Li, Armstrong, Agarwal, Liu, Finn, and Manning}]{mitchell-etal-2022-enhancing}
Eric Mitchell, Joseph Noh, Siyan Li, Will Armstrong, Ananth Agarwal, Patrick Liu, Chelsea Finn, and Christopher Manning. 2022.
\newblock \href {https://doi.org/10.18653/v1/2022.emnlp-main.115} {Enhancing self-consistency and performance of pre-trained language models through natural language inference}.
\newblock In \emph{Proceedings of the 2022 Conference on Empirical Methods in Natural Language Processing}, pages 1754--1768, Abu Dhabi, United Arab Emirates. Association for Computational Linguistics.

\bibitem[{Narayan et~al.(2018)Narayan, Cohen, and Lapata}]{narayan-etal-2018-dont}
Shashi Narayan, Shay~B. Cohen, and Mirella Lapata. 2018.
\newblock \href {https://doi.org/10.18653/v1/D18-1206} {Don{'}t give me the details, just the summary! topic-aware convolutional neural networks for extreme summarization}.
\newblock In \emph{Proceedings of the 2018 Conference on Empirical Methods in Natural Language Processing}, pages 1797--1807, Brussels, Belgium. Association for Computational Linguistics.

\bibitem[{Ning et~al.(2017)Ning, Feng, and Roth}]{ning-etal-2017-structured}
Qiang Ning, Zhili Feng, and Dan Roth. 2017.
\newblock \href {https://doi.org/10.18653/v1/D17-1108} {A structured learning approach to temporal relation extraction}.
\newblock In \emph{Proceedings of the 2017 Conference on Empirical Methods in Natural Language Processing}, pages 1027--1037, Copenhagen, Denmark. Association for Computational Linguistics.

\bibitem[{Ning et~al.(2018)Ning, Feng, Wu, and Roth}]{ning-etal-2018-joint}
Qiang Ning, Zhili Feng, Hao Wu, and Dan Roth. 2018.
\newblock \href {https://doi.org/10.18653/v1/P18-1212} {Joint reasoning for temporal and causal relations}.
\newblock In \emph{Proceedings of the 56th Annual Meeting of the Association for Computational Linguistics (Volume 1: Long Papers)}, pages 2278--2288, Melbourne, Australia. Association for Computational Linguistics.

\bibitem[{Ning et~al.(2020)Ning, Wu, Han, Peng, Gardner, and Roth}]{ning-etal-2020-torque}
Qiang Ning, Hao Wu, Rujun Han, Nanyun Peng, Matt Gardner, and Dan Roth. 2020.
\newblock \href {https://doi.org/10.18653/v1/2020.emnlp-main.88} {{TORQUE}: A reading comprehension dataset of temporal ordering questions}.
\newblock In \emph{Proceedings of the 2020 Conference on Empirical Methods in Natural Language Processing (EMNLP)}, pages 1158--1172, Online. Association for Computational Linguistics.

\bibitem[{Ouyang et~al.(2022)Ouyang, Wu, Jiang, Almeida, Wainwright, Mishkin, Zhang, Agarwal, Slama, Ray et~al.}]{ouyang2022training}
Long Ouyang, Jeffrey Wu, Xu~Jiang, Diogo Almeida, Carroll Wainwright, Pamela Mishkin, Chong Zhang, Sandhini Agarwal, Katarina Slama, Alex Ray, et~al. 2022.
\newblock Training language models to follow instructions with human feedback.
\newblock \emph{Advances in Neural Information Processing Systems}, 35:27730--27744.

\bibitem[{Penedo et~al.(2023)Penedo, Malartic, Hesslow, Cojocaru, Cappelli, Alobeidli, Pannier, Almazrouei, and Launay}]{refinedweb}
Guilherme Penedo, Quentin Malartic, Daniel Hesslow, Ruxandra Cojocaru, Alessandro Cappelli, Hamza Alobeidli, Baptiste Pannier, Ebtesam Almazrouei, and Julien Launay. 2023.
\newblock \href {http://arxiv.org/abs/2306.01116} {The {R}efined{W}eb dataset for {F}alcon {LLM}: outperforming curated corpora with web data, and web data only}.
\newblock \emph{arXiv preprint arXiv:2306.01116}.

\bibitem[{Qin et~al.(2022)Qin, Welleck, Khashabi, and Choi}]{qin2022cold}
Lianhui Qin, Sean Welleck, Daniel Khashabi, and Yejin Choi. 2022.
\newblock Cold decoding: Energy-based constrained text generation with langevin dynamics.
\newblock \emph{Advances in Neural Information Processing Systems}, 35:9538--9551.

\bibitem[{Qin et~al.(2024)Qin, Song, Hu, Yao, Cho, Wang, Wu, Liu, Liu, and Yu}]{qin2024infobench}
Yiwei Qin, Kaiqiang Song, Yebowen Hu, Wenlin Yao, Sangwoo Cho, Xiaoyang Wang, Xuansheng Wu, Fei Liu, Pengfei Liu, and Dong Yu. 2024.
\newblock Infobench: Evaluating instruction following ability in large language models.
\newblock \emph{arXiv preprint arXiv:2401.03601}.

\bibitem[{Rafailov et~al.(2023)Rafailov, Sharma, Mitchell, Ermon, Manning, and Finn}]{rafailov2023direct}
Rafael Rafailov, Archit Sharma, Eric Mitchell, Stefano Ermon, Christopher~D Manning, and Chelsea Finn. 2023.
\newblock Direct preference optimization: Your language model is secretly a reward model.
\newblock \emph{arXiv preprint arXiv:2305.18290}.

\bibitem[{Robinson et~al.(2021)Robinson, Chuang, Sra, and Jegelka}]{robinson2021contrastive}
Joshua Robinson, Ching-Yao Chuang, Suvrit Sra, and Stefanie Jegelka. 2021.
\newblock Contrastive learning with hard negative samples.
\newblock In \emph{International Conference on Learning Representations (ICLR)}.

\bibitem[{Roth and Yih(2004)}]{roth2004linear}
Dan Roth and Wen-tau Yih. 2004.
\newblock A linear programming formulation for global inference in natural language tasks.
\newblock In \emph{Proceedings of the eighth conference on computational natural language learning (CoNLL-2004) at HLT-NAACL 2004}, pages 1--8.

\bibitem[{Sun et~al.(2023)Sun, Tian, Zhou, Xu, Hu, Gupta, Wieting, Peng, and Ma}]{sun2023evaluating}
Jiao Sun, Yufei Tian, Wangchunshu Zhou, Nan Xu, Qian Hu, Rahul Gupta, John~Frederick Wieting, Nanyun Peng, and Xuezhe Ma. 2023.
\newblock Evaluating large language models on controlled generation tasks.
\newblock \emph{arXiv preprint arXiv:2310.14542}.

\bibitem[{Taori et~al.(2023)Taori, Gulrajani, Zhang, Dubois, Li, Guestrin, Liang, and Hashimoto}]{taori2023alpaca}
Rohan Taori, Ishaan Gulrajani, Tianyi Zhang, Yann Dubois, Xuechen Li, Carlos Guestrin, Percy Liang, and Tatsunori~B Hashimoto. 2023.
\newblock Alpaca: A strong, replicable instruction-following model.
\newblock \emph{Stanford Center for Research on Foundation Models. https://crfm. stanford. edu/2023/03/13/alpaca. html}, 3(6):7.

\bibitem[{Vijayakumar et~al.(2018)Vijayakumar, Cogswell, Selvaraju, Sun, Lee, Crandall, and Batra}]{vijayakumar2018diverse}
Ashwin Vijayakumar, Michael Cogswell, Ramprasaath Selvaraju, Qing Sun, Stefan Lee, David Crandall, and Dhruv Batra. 2018.
\newblock Diverse beam search for improved description of complex scenes.
\newblock In \emph{Proceedings of the AAAI Conference on Artificial Intelligence}, volume~32.

\bibitem[{Walker et~al.(2006)Walker, Strassel, Medero, and Maeda}]{walker2006ace}
Christopher Walker, Stephanie Strassel, Julie Medero, and Kazuaki Maeda. 2006.
\newblock Ace 2005 multilingual training corpus.
\newblock \emph{Linguistic Data Consortium, Philadelphia}, 57:45.

\bibitem[{Wang et~al.(2022{\natexlab{a}})Wang, Song, Zhang, Jin, Cho, Yao, Wang, Chen, and Yu}]{wang-etal-2022-salience}
Fei Wang, Kaiqiang Song, Hongming Zhang, Lifeng Jin, Sangwoo Cho, Wenlin Yao, Xiaoyang Wang, Muhao Chen, and Dong Yu. 2022{\natexlab{a}}.
\newblock \href {https://doi.org/10.18653/v1/2022.emnlp-main.409} {Salience allocation as guidance for abstractive summarization}.
\newblock In \emph{Proceedings of the 2022 Conference on Empirical Methods in Natural Language Processing}, pages 6094--6106, Abu Dhabi, United Arab Emirates. Association for Computational Linguistics.

\bibitem[{Wang et~al.(2020)Wang, Chen, Zhang, and Roth}]{wang-etal-2020-joint}
Haoyu Wang, Muhao Chen, Hongming Zhang, and Dan Roth. 2020.
\newblock \href {https://doi.org/10.18653/v1/2020.emnlp-main.51} {Joint constrained learning for event-event relation extraction}.
\newblock In \emph{Proceedings of the 2020 Conference on Empirical Methods in Natural Language Processing (EMNLP)}, pages 696--706, Online. Association for Computational Linguistics.

\bibitem[{Wang et~al.(2023{\natexlab{a}})Wang, He, Nguyen, Kumar, and Roth}]{wang2023regularization}
Kaifu Wang, Hangfeng He, Tin~D Nguyen, Piyush Kumar, and Dan Roth. 2023{\natexlab{a}}.
\newblock On regularization and inference with label constraints.
\newblock \emph{Proceedings of the 40 th International Conference on Machine Learning}.

\bibitem[{Wang et~al.(2023{\natexlab{b}})Wang, Kordi, Mishra, Liu, Smith, Khashabi, and Hajishirzi}]{wang-etal-2023-self-instruct}
Yizhong Wang, Yeganeh Kordi, Swaroop Mishra, Alisa Liu, Noah~A. Smith, Daniel Khashabi, and Hannaneh Hajishirzi. 2023{\natexlab{b}}.
\newblock \href {https://doi.org/10.18653/v1/2023.acl-long.754} {Self-instruct: Aligning language models with self-generated instructions}.
\newblock In \emph{Proceedings of the 61st Annual Meeting of the Association for Computational Linguistics (Volume 1: Long Papers)}, pages 13484--13508, Toronto, Canada. Association for Computational Linguistics.

\bibitem[{Wang et~al.(2022{\natexlab{b}})Wang, Mishra, Alipoormolabashi, Kordi, Mirzaei, Naik, Ashok, Dhanasekaran, Arunkumar, Stap, Pathak, Karamanolakis, Lai, Purohit, Mondal, Anderson, Kuznia, Doshi, Pal, Patel, Moradshahi, Parmar, Purohit, Varshney, Kaza, Verma, Puri, Karia, Doshi, Sampat, Mishra, Reddy~A, Patro, Dixit, and Shen}]{wang-etal-2022-super}
Yizhong Wang, Swaroop Mishra, Pegah Alipoormolabashi, Yeganeh Kordi, Amirreza Mirzaei, Atharva Naik, Arjun Ashok, Arut~Selvan Dhanasekaran, Anjana Arunkumar, David Stap, Eshaan Pathak, Giannis Karamanolakis, Haizhi Lai, Ishan Purohit, Ishani Mondal, Jacob Anderson, Kirby Kuznia, Krima Doshi, Kuntal~Kumar Pal, Maitreya Patel, Mehrad Moradshahi, Mihir Parmar, Mirali Purohit, Neeraj Varshney, Phani~Rohitha Kaza, Pulkit Verma, Ravsehaj~Singh Puri, Rushang Karia, Savan Doshi, Shailaja~Keyur Sampat, Siddhartha Mishra, Sujan Reddy~A, Sumanta Patro, Tanay Dixit, and Xudong Shen. 2022{\natexlab{b}}.
\newblock \href {https://doi.org/10.18653/v1/2022.emnlp-main.340} {Super-{N}atural{I}nstructions: Generalization via declarative instructions on 1600+ {NLP} tasks}.
\newblock In \emph{Proceedings of the 2022 Conference on Empirical Methods in Natural Language Processing}, pages 5085--5109, Abu Dhabi, United Arab Emirates. Association for Computational Linguistics.

\bibitem[{Yuan et~al.(2023)Yuan, Yuan, Tan, Wang, Huang, and Huang}]{yuan2023rrhf}
Zheng Yuan, Hongyi Yuan, Chuanqi Tan, Wei Wang, Songfang Huang, and Fei Huang. 2023.
\newblock Rrhf: Rank responses to align language models with human feedback without tears.
\newblock \emph{arXiv preprint arXiv:2304.05302}.

\bibitem[{Zhang et~al.(2023{\natexlab{a}})Zhang, Dong, Li, Zhang, Sun, Wang, Li, Hu, Zhang, Wu et~al.}]{zhang2023instruction}
Shengyu Zhang, Linfeng Dong, Xiaoya Li, Sen Zhang, Xiaofei Sun, Shuhe Wang, Jiwei Li, Runyi Hu, Tianwei Zhang, Fei Wu, et~al. 2023{\natexlab{a}}.
\newblock Instruction tuning for large language models: A survey.
\newblock \emph{arXiv preprint arXiv:2308.10792}.

\bibitem[{Zhang et~al.(2019)Zhang, Kishore, Wu, Weinberger, and Artzi}]{zhang2019bertscore}
Tianyi Zhang, Varsha Kishore, Felix Wu, Kilian~Q Weinberger, and Yoav Artzi. 2019.
\newblock Bertscore: Evaluating text generation with bert.
\newblock In \emph{International Conference on Learning Representations}.

\bibitem[{Zhang et~al.(2023{\natexlab{b}})Zhang, Ladhak, Durmus, Liang, McKeown, and Hashimoto}]{zhang2023benchmarking}
Tianyi Zhang, Faisal Ladhak, Esin Durmus, Percy Liang, Kathleen McKeown, and Tatsunori~B Hashimoto. 2023{\natexlab{b}}.
\newblock Benchmarking large language models for news summarization.
\newblock \emph{arXiv preprint arXiv:2301.13848}.

\bibitem[{Zhou et~al.(2023)Zhou, Liu, Xu, Iyer, Sun, Mao, Ma, Efrat, Yu, Yu et~al.}]{zhou2023lima}
Chunting Zhou, Pengfei Liu, Puxin Xu, Srini Iyer, Jiao Sun, Yuning Mao, Xuezhe Ma, Avia Efrat, Ping Yu, Lili Yu, et~al. 2023.
\newblock Lima: Less is more for alignment.
\newblock \emph{arXiv preprint arXiv:2305.11206}.

\bibitem[{Zhu et~al.(2021)Zhu, Hinthorn, Xu, Zeng, Zeng, Huang, and Jiang}]{zhu-etal-2021-enhancing}
Chenguang Zhu, William Hinthorn, Ruochen Xu, Qingkai Zeng, Michael Zeng, Xuedong Huang, and Meng Jiang. 2021.
\newblock \href {https://doi.org/10.18653/v1/2021.naacl-main.58} {Enhancing factual consistency of abstractive summarization}.
\newblock In \emph{Proceedings of the 2021 Conference of the North American Chapter of the Association for Computational Linguistics: Human Language Technologies}, pages 718--733, Online. Association for Computational Linguistics.

\end{thebibliography}
